%% file: ms.tex
\documentclass[a4paper]{article}
\usepackage[letterpaper,top=3cm,bottom=2cm,left=2cm,right=1.5cm,marginparwidth=1.75cm]{geometry}

\usepackage[authoryear,super,sort&compress,comma]{natbib} 
\usepackage{authblk}

\usepackage{microtype}
\usepackage{graphicx}
\usepackage{subfigure}
\usepackage{booktabs}
\usepackage{hyperref}
\usepackage{amsmath}
\usepackage{amssymb}
\usepackage{mathtools}
\usepackage{amsthm}
\usepackage[capitalize,noabbrev]{cleveref}

\input{math_commands.tex}

\usepackage[textsize=tiny]{todonotes}

\title{Symmetry-Aware Bayesian Flow Networks for Crystal Generation}
\author[1]{Laura Ruple}
\author[1,2]{Luca Torresi}
\author[1]{Henrik Schopmans}
\author[1,2]{Pascal Friederich*}
\affil[1]{Institute of Theoretical Informatics, Karlsruhe Institute of Technology, Kaiserstr. 12, 76131 Karlsruhe, Germany}
\affil[2]{Institute of Nanotechnology, Karlsruhe Institute of Technology, Kaiserstr. 12, 76131 Karlsruhe, Germany}
\affil[*]{Corresponding author: pascal.friederich@kit.edu}

\renewcommand{\cite}{\citep}

\begin{document}

\twocolumn[
\maketitle
]

\begin{abstract}
The discovery of new crystalline materials is essential to scientific and
technological progress. However, traditional trial-and-error approaches are
inefficient due to the vast search space. Recent advancements in machine
learning have enabled generative models to predict new stable materials by
incorporating structural symmetries and to condition the generation on desired
properties. In this work, we introduce SymmBFN, a novel symmetry-aware Bayesian
Flow Network (BFN) for crystalline material generation that accurately
reproduces the distribution of space groups found in experimentally observed
crystals. SymmBFN substantially improves efficiency, generating stable
structures at least 50 times faster than the next-best method. Furthermore, we
demonstrate its capability for property-conditioned generation, enabling the
design of materials with tailored properties. Our findings establish BFNs as an
effective tool for accelerating the discovery of crystalline materials.
\end{abstract}

\section{Introduction}
The discovery and design of novel materials are essential for advancing
technologies in areas such as energy, electronics, and sustainability. Recent
developments in machine learning are reshaping the field of materials research
in diverse and transformative ways \cite{Mobarak_genMLreview}. Machine learning
models have been applied to enhance conventional screening approaches
\cite{Gnome, schmidtLargescaleMachinelearningassistedExploration2022}. At the
same time, advancements in generative modelling show great potential for
accelerating materials design by proposing novel and realistic candidates for
crystal structures with targeted properties \cite{Park2024_genai_mat_review}. \\
An early demonstration of the effectiveness of generative models for crystalline
materials is CDVAE \cite{cdvae}, which employs a variational autoencoder to
generate composition, lattice parameters, and the number of atoms in the unit
cell. Then, the atom coordinates are randomly initialised and iteratively
denoised using score-matching \cite{song2019generative}. Subsequent works
\cite{diffcsp, zeni2023mattergen, UniMat} have replaced the autoencoder with a
purely diffusion-based approach, jointly diffusing lattice parameters,
fractional coordinates, and atom types. This approach effectively captures
crystal geometries as a whole, leading to an improved quality of the generated
structures. Various other generative modelling techniques have been applied to
crystals, including Riemannian flow matching \cite{flowmm},  large language
models \cite{crystallm_Antunes, crystallm_gruver, flowllm}, and normalizing
flows \cite{wirnsberger2022normalizing}. While these models have demonstrated
the capability to generate novel and stable structures, an important aspect that
has often been neglected is the incorporation of space group symmetry. A large
proportion of the samples generated with models such as DiffCSP \cite{diffcsp}
and CDVAE belong to the low-symmetry space group P1, which is rarely observed in
nature. To alleviate this issue, models that explicitly incorporate space group
symmetries into the generation process have been recently proposed
\cite{diffcsp++, symmcd}. \\
Diffusion-based methods have proven effective in predicting realistic materials.
However, they present two main limitations when applied to crystal generation:
The first issue is the lack of a unified diffusion framework that can
accommodate different variable types. This poses a challenge due to the inherent
multi-modality of crystal structure representations, where, for example,
continuous variables are used for atomic coordinates and lattice parameters,
while categorical variables represent atom types and site symmetries. A second
drawback is that generating high-quality samples requires a significant number
of integration steps, which results in high computational costs. \\
Bayesian Flow Networks (BFN) are a novel class of generative models proposed by
\citet{graves2023bayesian}, which generate samples through an iterative
procedure similar to the reverse process used in diffusion. Unlike diffusion
models, BFNs operate directly on the parameters of the data distribution, rather
than its noisy samples. This approach allows for a unified framework capable of
uniformly handling categorical, discrete, and continuous variables. BFNs have
been recently applied to the domain of three-dimensional molecular generation
\cite{song2023unified}, where they demonstrated a superior trade-off between
efficiency and quality, achieving a significant speedup over diffusion models.
\\
\textbf{Contribution:} In this study, we present SymmBFN, a novel adaptation of
BFNs for the generation of crystalline materials using a symmetry-aware
representation of crystal structures first introduced by \citet{symmcd}. The
proposed method addresses several limitations of previous models, by i) enabling
the joint modelling of distributions over fractional coordinates, atom types,
unit cell parameters, and site symmetries within a unified framework, ii)
accurately reproducing the distribution of space groups observed in real-world
materials, iii) achieving a speed improvement in the generation process of up to
two orders of magnitude compared to the state-of-the-art. Additionally, we
develop a method to condition the generation process on additional desired
properties. Our results demonstrate the potential of BFNs as a tool for
accelerating the design of crystalline materials with targeted properties.

\section{Preliminaries}
\subsection{Representation of Crystal Structures}
\label{sec:unitcell}
\textbf{Unit cell:} Crystal structures are represented by the unit cell
\cite{crystals_basics}, which is a repeating unit that describes the arrangement
of atoms in a crystal. A unit cell $\C= (\lattice, \A, \x)$ is defined by the
lattice $\lattice = \left(\didx{\mathbf{l}}{1}, \didx{\mathbf{l}}{2},
\didx{\mathbf{l}}{3}\right) \in \mathbb{R}^{3\times3}$ , the atomic numbers
$\A = \left(\didx{a}{1},\dots, \didx{a}{D}\right) \in \dsd{K}{D}$ of the ${D}$
atoms it contains, and their fractional coordinates $\x =
\left(\didx{\x}{1},\dots, \didx{\x}{D}\right)\in [0,1)^{3\times D}$. The
fractional coordinates $\x$ express the positions of the atoms relative to the
lattice vectors, corresponding to the Cartesian coordinates ${\Tilde{\x} =
\sum_{i=1}^{3}x_i\mathbf{l}_i \in \mathbb{R}^{3}}$. In an ideal crystal
structure, this arrangement repeats itself infinitely in all three dimensions.
The periodicity of the fractional coordinates is captured by the equivalence
relation ${\x \sim \x + \mathbf{m}\, , \text{with }\mathbf{m} \in \mathbb{Z}^3
}$. The periodicity of the crystal structure ensures its invariance under
translations. Specifically, the structure remains unchanged when a translation
vector \( \mathbf{t} \in \mathbb{R}^3 \) is applied uniformly to the entire unit
cell. \label{sec:symmetries}\\
\textbf{Space groups and point groups:}  The symmetry of a crystal structure is
described by its space group $G$, a subgroup of the Euclidean group
\cite{crystal_symmetries}. It contains all Euclidean transformations
$(\mathbf{O}, \mathbf{t})$, with $\mathbf{O} \in O(n)$ and $\mathbf{t} \in
\mathbb{R}^3$, that leave the crystal unchanged under the action $(\mathbf{O},
\mathbf{t})\x = \mathbf{O}\x + \mathbf{t}$ for $\x \in \mathbb{R}^3$. Two space
groups are of the same type if there exists an orientation-preserving Euclidean
transformation that maps all operations of one space group to the other. For
three-dimensional crystal structures, there are 230 different types of space
groups in total, representing different combinations of reflections, rotations,
inversions, translations, screw axes, and glide planes. The point group $P$ of a
crystal structure is the set of operations that leave at least one point in the
structure unchanged. The point group type is defined similarly to the space
group type, resulting in 32 different crystallographic point groups. \\
\textbf{Site symmetries and Wyckoff positions:} Given the space group $G$ of a
crystal structure, we can describe the symmetries of the atoms it is composed
of. The site symmetry group $S_{\x}$ of a point $\x$ in the unit cell is defined
as the subgroup of space group transformations that leave $\x$ invariant:
$S_{\x} = \left\{(\mathbf{O}, \mathbf{t}) \in G \mid  (\mathbf{O}, \mathbf{t})\x
= \x\right\}$. Thus, the site symmetry groups of the atoms in a crystal
structure are subgroups of the point group $P$ of the structure.
\begin{figure}[ht] 
\vskip 0.2in
\begin{center}
\centerline{\includegraphics[width=0.6\columnwidth]{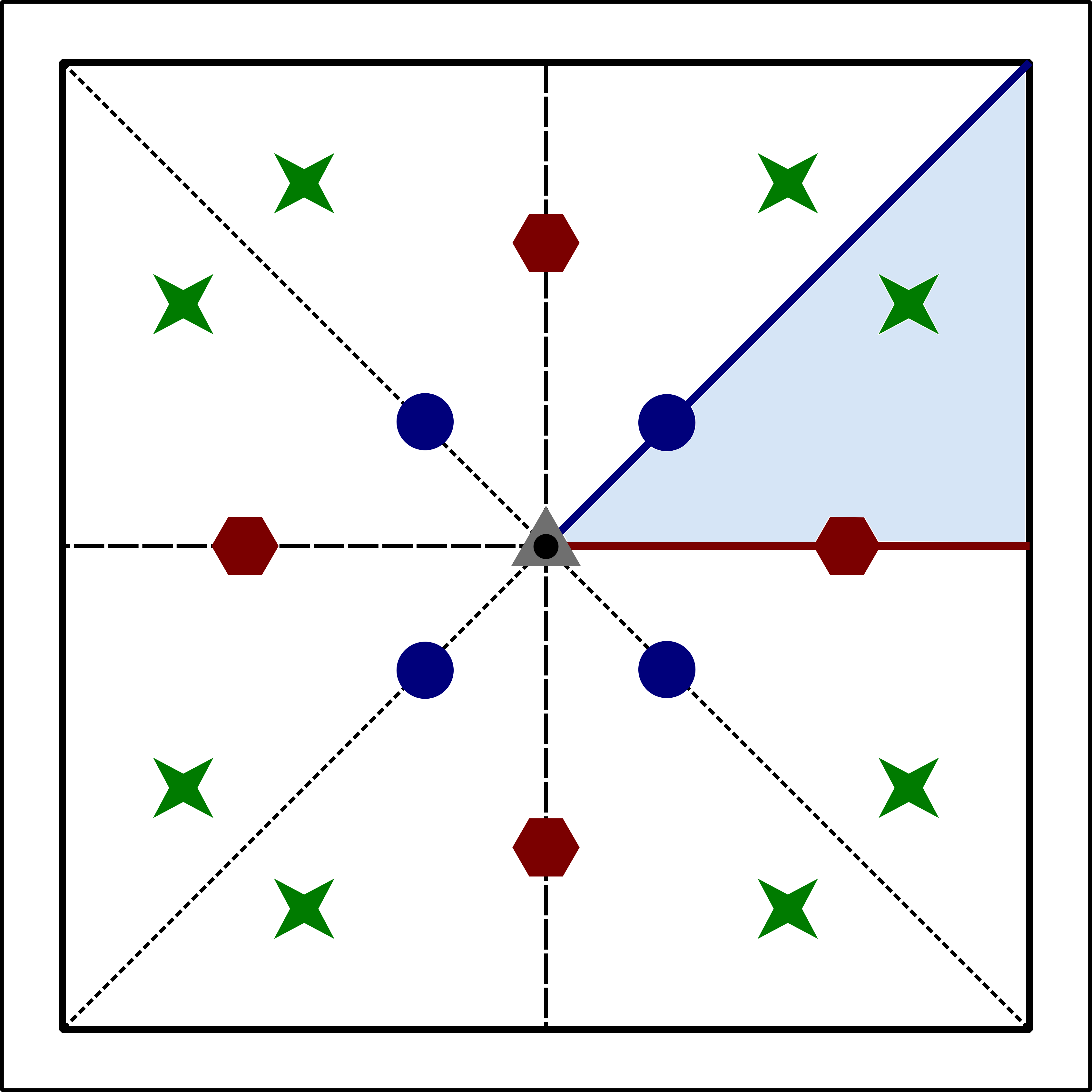}}
\caption{\textbf{Visualisation of the 2D crystallographic point group 4mm}. The
symbols indicate different Wyckoff positions and the dashed lines show the
symmetry axes. The light blue triangle is the asymmetric unit.}
\label{fig:wyckoff}
\end{center}
\vskip -0.2in
\end{figure} 
Further, the points $\x$ and $\mathbf{x'}$ are part of the same crystallographic
orbit if there is a transformation $(\mathbf{O}, \mathbf{t}) \in G $ with $
(\mathbf{O}, \mathbf{t})\x = \mathbf{x'}$. If the orbits of points $\x$ and
$\mathbf{x'}$ have conjugate site symmetry groups $S_{\x}$ and $S_\mathbf{x'}$,
they belong to the same class of crystallographic orbit, which is called a
Wyckoff position. Two  site symmetry groups $S_{\x}$ and $S_\mathbf{x'}$ are
conjugate, if there is an operation $g \in G$ such that $S_{\x'} =
g^{-1}S_{\x}g$. There are general Wyckoff positions whose site symmetry groups
contain only the identity operation, and special Wyckoff positions which are
left invariant by at least one other transformation of the space group. The
multiplicity of the Wyckoff positions of point $\x$ expresses the number of
points that occupy the same position and is given by $ \mathopen\mid
P\mathopen\mid /\mathopen\mid S_{\x} \mathopen\mid$. In a crystal structure,
there are 15 different possible symmetry axes \cite{tables_crystallography,
symmcd}, including the base, body, and face diagonals. The space group
determines which of 13 possible symmetry operations, such as rotation or
roto-inversion, can occur on each axis. Therefore, the site symmetry groups of
each Wyckoff position can be fully described by specifying the corresponding
symmetry operation for each axis. \\
\textbf{Asymmetric cell:} Using the symmetries of the crystal, the whole
structure can be described by the asymmetric unit, which contains only one
representative for each class of crystallographic orbit. The full structure can
then be restored by applying the symmetry operations of the space group. As an
example, Figure~\ref{fig:wyckoff} shows a visualisation of the 2D
crystallographic point group 4mm \cite{Bilbao_Crystallographic_Server}. The
coloured symbols indicate the occupied positions in the structure, while the
dashed lines are the symmetry axes. The grey triangle is placed in the position
with the highest symmetry. Thus, it is the only representative of its Wyckoff
position. Its site symmetry group is equivalent to the whole point group of the
structure. The blue circle and red hexagon each have a multiplicity of 4. The
site symmetry group for the blue circle includes the identity and mirror
symmetries along the diagonals, while for the red hexagon, it includes the
identity and mirror symmetries along the centre lines. The green stars occupy
the general Wyckoff position, as their site symmetry group contains only the
identity operation. The light blue triangle on the top right shows the
asymmetric unit. The entire cell can be reconstructed by applying all the
symmetry operations of the point group to the four representative elements.
\begin{table}[ht]
\caption{Constraints on $k_i$ for each space group \cite{diffcsp++}.}
\vskip 0.15in
\begin{center}
\begin{small}
\begin{sc}
\centering
\begin{tabular}{cc}
\toprule
   Space Group & Constraints on $k_i$ \\
\midrule\midrule
    $1-2$  & $(k_1,k_2,k_3,k_4,k_5,k_6)$ \\\midrule
     $3-15$   & $(0,k_2,0,k_4,k_5,k_6)$ \\\midrule
     $16-74$   & $(0,0,0,k_4,k_5,k_6)$ \\\midrule
    {$75-142$}  & $(0,0,0,0,k_5,k_6)$ \\\midrule
    {$143-194$}  & $(-log(3)/4,0,0,0,k_5,k_6)$ \\\midrule
    {$195-230$}   & $(0,0,0,0,0,k_6)$\\
\bottomrule
\end{tabular}
\end{sc}
\end{small}
\end{center}
\vskip -0.1in

\label{tab:k_cons}
\end{table}

\textbf{Constraints on the lattice:} The symmetries defined by the space group
of a crystal structure impose constraints on the shape of the lattice
$\lattice$. To comply with these constraints, \citet{diffcsp++} developed the
following notation for the lattice. First, the lattice is decomposed into
$\lattice=\mQ\exp(\mS)$, where $\mQ\in\mathbb{R}^{3\times 3}$ is an orthogonal
matrix and $\mS\in\mathbb{R}^{3\times 3}$ is a symmetric matrix. Thus,
$\lattice$ can be completely defined by $\mS$, while $\mS$ is not affected by
any orthogonal transformation on $\lattice$, as these only change $\mQ$.
\citeauthor{diffcsp++} use the basis $\mB$ for the subspace of symmetric
matrices such that ${\mS=\sum_{i=1}^6 k_i\mB_i, k_i\in \mathbb{R}}$, where the
matrices $\mB_i$ are defined as 
\begin{align}
\begin{aligned}
\mB_1 &= \begin{pmatrix} 0 & 1 & 0 \\ 1 & 0 & 0 \\ 0 & 0 & 0 \end{pmatrix}\, , &
\mB_2 &= \begin{pmatrix} 0 & 0 & 1 \\ 0 & 0 & 0 \\ 1 & 0 & 0 \end{pmatrix}\, , \\
\mB_3 &= \begin{pmatrix} 0 & 0 & 0 \\ 0 & 0 & 1 \\ 0 & 1 & 0 \end{pmatrix}\, , &
\mB_4 &= \begin{pmatrix} 1 & 0 & 0 \\ 0 & -1 & 0 \\ 0 & 0 & 0 \end{pmatrix}\, , \\
\mB_5 &= \begin{pmatrix} 1 & 0 & 0 \\ 0 & 1 & 0 \\ 0 & 0 & -2 \end{pmatrix}\, , &
\mB_6 &= \begin{pmatrix} 1 & 0 & 0 \\ 0 & 1 & 0 \\ 0 & 0 & 1 \end{pmatrix}\, .
\end{aligned}
\end{align}

This way, the lattice $\lattice$ is fully determined by the values $k_i$. The
space group imposes constraints on the choice of the values $k_i$ (see Table
\ref{tab:k_cons}).

\subsection{Bayesian Flow Networks}
\begin{figure}[ht] 
\vskip 0.2in
\begin{center}
\centerline{\includegraphics[width=0.9\columnwidth]{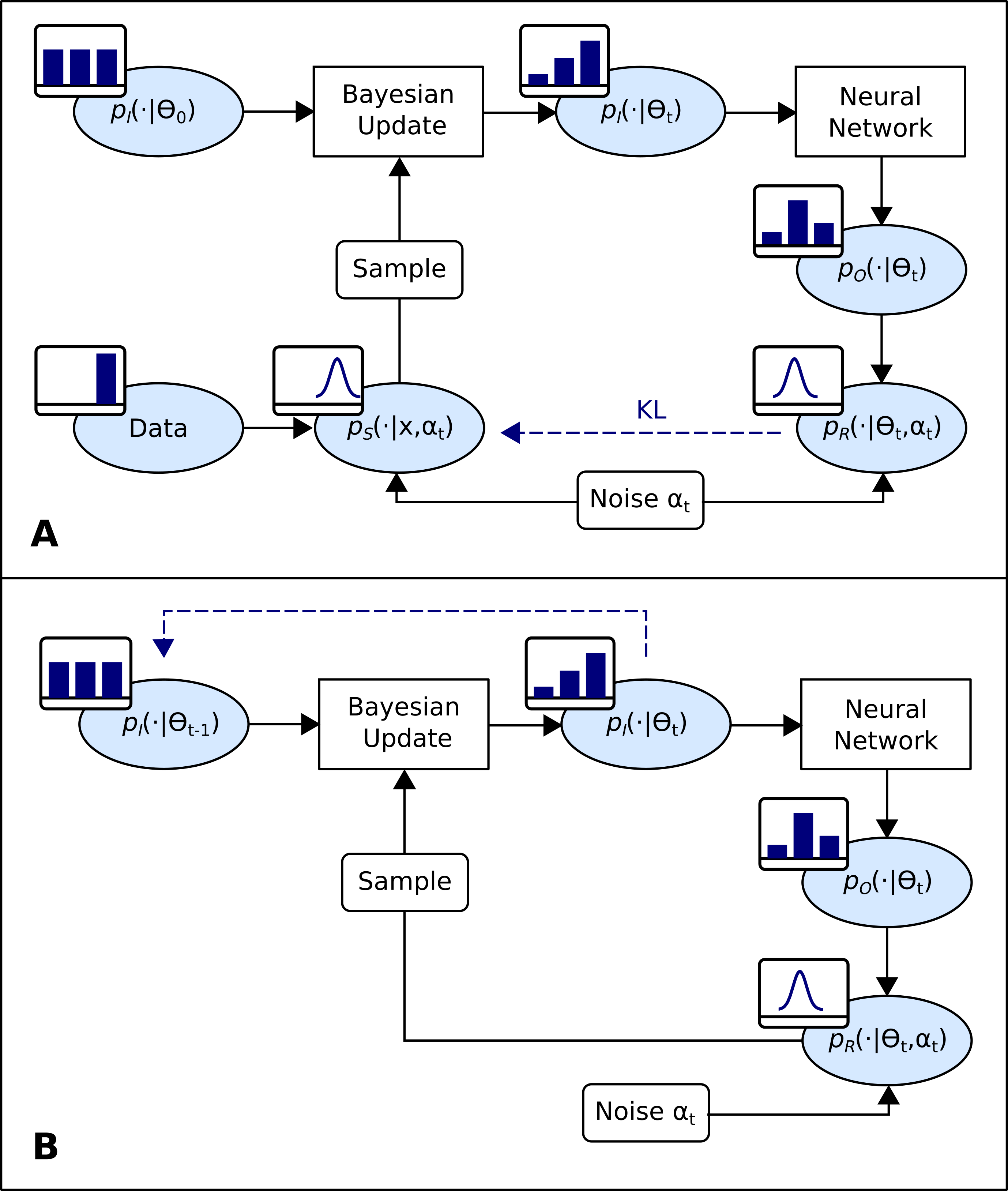}}
\caption{\textbf{Bayesian Flow Network.} Illustration of BFN training procedure
at the upper panel (A) and of the sampling procedure at the lower panel (B). }
\label{fig:overview}
\end{center}
\vskip -0.2in
\end{figure}

BFNs \cite{graves2023bayesian} generate data through an iterative process that
combines the strengths of Bayesian inference and deep learning. This process
(see Figure~\ref{fig:overview}) involves transforming the parameters of a
distribution representing the data from an uninformative prior to increasingly
confident posteriors, alternating between two steps: i) The parameters of a set
of independent distributions $\inp $, each representing a variable, are updated
using Bayesian inference. ii) These updated parameters are passed to a neural
network, incorporating contextual information to output a joint distribution
$\out$.\\
Given $D$-dimensional data $\x = \left(\didx{x}{1},\dots,\didx{x}{D}\right) \in
\X^D$, the factorised \emph{input distribution} is defined as 
\begin{align}
\inp(\x \mid \parsn) = \prod_{d=1}^D \inp(\didx{x}{d} \mid \parsdd{d})\, .
\end{align}

During training, samples are drawn from the \emph{sender distribution},
generated by adding random noise to the data:
\begin{align}
\sender{\y}{\x;\alpha} = \prod_{d=1}^D \sender{\didx{y}{d}}{\didx{x}{d}; \alpha}\, .
\end{align}
The \emph{accuracy} parameter $\alpha \in \R^+$ controls the noise level and is
adjusted according to a predefined schedule, dependent on process \emph{time} $t
\in [0,1]$, to control the informativeness of the samples at different stages of
the process. Given a sample $\y$, the parameters of the input distribution can
be updated through a \emph{Bayesian update function} ${h: \parsn' \leftarrow
h(\parsn, \y, \alpha)}$.
\label{sec:bfn}
The \emph{Bayesian update distribution} is thus obtained by marginalizing out $\y$:
\begin{align}
\update(\parsn' \mid \parsn, \x; \alpha) = \E_{\sender{\y}{\x;\alpha}} \delta \left(\parsn' -h(\parsn, \y, \alpha) \right)\, . 
\label{update}
\end{align}
To capture the interdependencies between different dimensions, the parameters
$\parsn$ and the process time $t$ are then fed into a neural network $\Psi$,
which outputs the parameters $\didx{\Psi}{d}(\parsn, t)$ of the \emph{output
distribution}
\begin{align}
\out(\x \mid \parsn, t) = \prod_{d=1}^D \out(\didx{x}{d} \mid \didx{\Psi}{d}(\parsn, t))\, .
\end{align}
Given the \emph{sender distribution} and the \emph{output distribution}, the
\emph{receiver distribution} is defined as
\begin{align}
\rec(\y \mid \parsn; t, \alpha) &= \E_{\out(\x' \mid \parsn; t)}\sender{\y}{\x'; \alpha}\, .
\label{r_dist}
\end{align}
For generating data, the \emph{receiver distribution} is used instead of the
\emph{sender distribution} to obtain the noisy samples required for the Bayesian
update. Finally, the loss function to train the neural network is given by the
negative variational lower bound
\begin{align}
L(\x)  &= \kl{q}{p} - \E_{\yt{1},\dots,\yt{n} \sim q} \ln \out(\x \mid \parsnt{n}; 1)\, ,
\label{vae_loss}
\end{align}
where
\begin{align}
&\left\{
\begin{aligned}
q(\yt{1},\dots,\yt{n}) &= \prod_{i=1}^n \sender{\y_i}{\x; \alpha_i}\, , \\
p(\yt{1},\dots,\yt{n}) &= \prod_{i=1}^n \rec(\y_i \mid \parsnt{i-1}; t_{i-1}, \alphat{i})\, .
\end{aligned}
\right.
\end{align}

\section{Methods}
SymmBFN employs BFNs to generate crystals constrained to any specified space
group, modelling in a single framework the lattice parameters, the site
symmetries, and the atoms in the asymmetric unit, and then reconstructing the
entire unit cell in a post-processing step. It is worth noting that since
SymmBFN utilizes a canonical reference system for crystal structures, namely the
unit cell axes, there is no need to enforce equivariance within the neural
network \cite{symmcd}.\\
This section details how each part of the crystal structure—namely the
fractional coordinates, atom types, lattice parameters, and site symmetry
groups—is handled within the SymmBFN framework. We then describe the
implementation of the neural network used to predict the parameters of the
output distribution, as well as the sampling process to generate new crystalline
structures. Finally, in Section \ref{sec:prop_gen}, we present an extension of
the model that allows for generation conditioned on additional desired
properties.\\
\textbf{Fractional coordinates:} We apply the BFN instantiation for continuous
data on the fractional coordinates $\x$ of the atoms in the asymmetric unit. The
\emph{input distribution} is set as an isotropic Gaussian ${\inp(\x \mid
\parsn^{\x}) = \N{\x \mid \m}{\rho^{-1}\I{D}}}$ with parameters ${\parsn^{\x} =
\{\m, \rho\}}$, and the prior in ${t=0}$ is initialized as ${\parsnt{0}^{\x} =
\{\0{D}, 1\}}$. Similarly, the \emph{sender distribution} takes the form of an
isotropic Gaussian: ${\sender{\y}{\x; \alpha} = \N{\y \mid
\x}{\alpha^{-1}\I{D}}.}$

Assuming both the \emph{input} and \emph{sender distribution} are isotropic
Gaussians, \citeauthor{graves2023bayesian} derive the \emph{Bayesian update
function} \emph{h} as
\begin{align}
&\left\{
\begin{aligned}
\pt{i} &= \pt{i-1} + \alpha\, , \\
\mt{i} &= \frac{\mt{i-1} \pt{i-1} + \y \alpha}{\pt{i}}\, .
\end{aligned}
\right.
\end{align}

Marginalizing over $\y \sim \N{\y \mid \x}{\alpha^{-1}\I{D}}$ yields the
\emph{Bayesian update distribution} for continuous data

\begin{align}
\update(\parsnt{i} \mid \parsnt{i-1}, \x; \alpha) = \N{\mt{i} \mid \frac{\alpha \x + \mt{i-1}\pt{i-1}}{\pt{i}}}{\frac{\alpha}{\pt{i
}^2}\I{D}}\, .
\label{cts_update_dist}
\end{align}

The \emph{Bayesian update} can be extended to continuous time by introducing the
accuracy schedule 

\begin{align}
\beta(t) = \int_{t'=0}^{t} \alpha(t') dt'\,.
\end{align}

The accuracy schedule for continuous data is defined as $\beta(t) =
\sigma_{\x}^{-2t} - 1$, where $\sigma_{\x}$ is the empirically chosen standard
deviation of the input distribution at $t=1$. \\
With ${\gamma(t) = \frac{\beta(t)}{1+\beta(t)} = 1 - \sigma_{\x}^{2t}}$, we
define the \emph{Bayesian flow distribution} as ${\flow(\parsn^{\x} \mid \x; t)
= \update(\parsn^{\x} \mid \parsnt{0}^{\x}, \x, \beta(t))}$, which for
continuous data can be derived to be 
\begin{align}
\flow(\parsn^{\x} \mid \x; t) &= \N{\m \mid \gamma(t)\x}{\gamma(t)(1-\gamma(t))\I{D}}\, .
\end{align}
The neural network is trained to predict an estimate $\eps(\parsn^{\x}, t)$ of
the Gaussian noise vector $\vec{\epsilon} \sim \N{\0{D}}{\I{D}}$ which was used
to generate the input $\m$ provided to the network. Given this estimate, the
\emph{output distribution} is 
\begin{align}
\out(\x \mid \parsn^{\x}; t) = \delta(\x-\mathbf{\pred{x}}(\parsn^{\x}, t))\label{cts_p_dist}\, ,
\end{align}
where
\begin{align}
\mathbf{\pred{x}}(\parsn^{\x}, t) = \frac{\m}{\gamma(t)} - \sqrt{\frac{1-\gamma(t)}{\gamma(t)}}\eps(\parsn^{\x}, t)\, .
\end{align}
The estimates $\mathbf{\pred{x}}(\parsn^{\x}, t)$ of the fractional coordinates
are then wrapped around the interval $[0,1)^3$ by applying the modulus
operation. Finally, the loss function is obtained as
\begin{align}
L^{\infty}(\x) &= -\ln \sigma_{\x} \E_{\substack{t \sim U(0,1), \\ \flow(\parsn^{\x} \mid \x; t)}}  \frac{\left\|\x - \mathbf{\pred{x}}(\parsn^{\x}, t)\right\|^2}{\sigma_{\x}^{2t}}\, .
\end{align}
 
\textbf{Atom types:} Consistently with the categorical nature of atom types, we
apply the BFN instantiation designed for discrete data. We represent the atom
type of each of the $D$ atoms in the asymmetric cell as $\A =
\left(\didx{a}{1},\dots, \didx{a}{D}\right) \in \dsd{K}{D}$, with $K$ being the
highest atomic number in the dataset. \\
The \emph{input distribution} for the atom types is defined as the categorical
distribution ${\inp(\A \mid \parsn^{\A}) = \prod_{d=1}^D
\pars_{\didx{a}{d}}^{(d)}}$ with parameters ${\parsn^{\A} =
\left(\pars_1^{(1)},\pars_2^{(1)},\dots,\pars_K^{(D)}\right) \in \R^{KD}}$,
where $\pars_k^{(d)}$ is the probability for atom $d$ to be of type $k$. The
prior is set to the uninformative uniform distribution, where all atom types are
assigned equal probability $1/K$. \\
The \emph{sender distribution} is then defined as 
\begin{align}
\sender{\y}{\A;\alpha} = \N{\y \mid \alpha\left(K \oh{\A}{KD} - \1{KD}\right)}{\alpha K \I{KD}}\label{disc_q_dist}\, ,
\end{align}
where ${\mathbf{e}_{\mathbf{a}} =
\left(\mathrm{e}^{(1)}_{1},\dots,\mathrm{e}^{(D)}_{K}\right) \in \R^{KD}}$ and
${\mathrm{e}^{(d)}_{k} = \delta_{k a^{(d)}}}$.\\
Substituting the \emph{Bayesian update function} for discrete data
${h(\parsnt{i-1}, \y, \alpha) = \frac{e^{\y}\parsnt{i-1}}{\sum_{k=1}^K
e^{\y_k}(\parsnt{i-1})_{k}}}$ into \eqref{update}, the \emph{Bayesian update
distribution} $\update(\parsn^{\A} \mid \parsnt{i-1}^{\A}, \A; \alpha)$ is given
by
\begin{align}
\ \E_{\N{\y \mid \alpha\left(K \oh{\A}{KD} - \1{KD}\right)}{\alpha K \I{KD}}} \delta\left(\parsn^{\A} - \frac{e^{\y}\parsnt{i-1}^{\A}}{\sum_{k=1}^K e^{\y_k}(\parsnt{i-1}^{\A})_{k}}\right)\, .
\end{align}
With the accuracy schedule defined as ${\beta(t) = t^2
\beta(1)\label{disc_beta_t}}$, where $\beta(1)$ is a hyperparameter, the
\emph{Bayesian flow distribution} for discrete data ${\flow(\parsn^{\A} \mid \A;
t)}$ is given by
\begin{align}
\E_{\substack {{\N{\y \mid \beta(t)\left(K \oh{\A}{KD} - \1{KD}\right)}{\beta(t) K \I{KD}}}}} \delta\left(\parsn^{\A} - \text{softmax}(\y)\right)\, .\label{disc_param_flow}
\end{align}

Given the network output $\eps_{\A} = \Psi(\parsn^{\A}, t)$, the \emph{output
distribution} can be obtained as
\begin{align}
\out(\A \mid \parsn^{\A}; t) = \prod_{d=1}^D \out^{(d)}(\Add{d} \mid \parsn^{\A}; t)\, ,\label{disc_pred_dist}
\end{align}
with
\begin{align}
\out^{(d)}(k \mid \parsn^{\A}; t) = \left(\text{softmax}(\eps_{\A}^{(d)})\right)_k\, .
\end{align}
Finally, the loss function for discrete data and continuous accuracy schedule
$\beta(t)$ is defined as
\begin{align}
L^{\infty}(\A) = K \beta(1) \E_{\substack {t\sim U(0,1), \\\flow(\parsn^{\A} \mid \A, t)}} t \|\oh{\A}{KD} - \mathbf{\pred{e}}(\parsn^{\A}, t)\|^2\, ,
\end{align}
where
\begin{align}
&\left\{
\begin{aligned}
&\mathbf{\pred{e}}(\parsn^{\A}, t) = \left(\mathbf{\pred{e}}^{(1)}(\parsn^{\A}, t),\dots,\mathbf{\pred{e}}^{(D)}(\parsn^{\A}, t)\right)\, , \\
&\mathbf{\pred{e}}^{(d)}(\parsn^{\A}, t) = \sum_{k=1}^K \out^{(d)}(k \mid \parsn^{\A}; t)\oh{k}{K}\, .
\end{aligned}
\right.
\end{align}

\textbf{Site symmetry groups:} For each atom in the asymmetric cell, we generate
the site symmetry group by having the model output the index of one of the 13
possible symmetry operations for each of the 15 axes. The model is trained with
the categorical BFN for the symmetry operations to generate $\mS = \left(
\didx{s}{1}, \didx{s}{2}, \dots, \didx{s}{15} \right) \in \{1,13\}^{15}$ for
each node. The BFN distributions for symmetry operations are defined similarly
to those for atomic numbers. The loss function for the symmetry operations is
thus defined as
\begin{align}
L^{\infty}(\mS) = 13 \cdot \beta_{\mS}(1)\E_{\substack {t\sim U(0,1), \\\flow(\parsn^{\mS} \mid \mS, t)}} t \|\oh{\mS}{KD} - \mathbf{\pred{e}}(\parsn^{\mS}, t)\|^2\, .
\end{align}
\textbf{Lattice:} We apply the BFN instantiation for continuous data on the
lattice vector representation $\mathbf{k} \in \mathbb{R}^6$, described in
Section \ref{sec:symmetries}. To comply with the space group constraint, after
the Bayesian updates and network calls, we introduce a masking step equivalent
to Table~\ref{tab:k_cons}. Otherwise, the distributions are defined analogously
to the fractional coordinate generation, with the loss function given by
\begin{align}
L^{\infty}(\mathbf{k}) &= -\ln \sigma_{\mathbf{k}} \E_{\substack{t \sim U(0,1)\, , \\ 
\flow(\parsn^{\mathbf{k}} \mid \mathbf{k}; t)}}  \frac{\left\|\mathbf{k} - \mathbf{\pred{\mathbf{k}}}(\parsn^{\mathbf{k}}, t)\right\|^2}{\sigma_{\mathbf{k}}^{2t}}\, .
\end{align}

\textbf{Neural network:} For our framework, we employ the graph neural network
architecture proposed by \citet{diffcsp}, based on the EGNN model
\cite{satorras2022egnn}. The network $\Psi(\m_{\k}, \m_{\x}, \parsn^{\A},
\parsn^{\mS}, t, G)$ predicts the scores for the output distributions after $N$
message-passing layers on a fully-connected graph. More details about the
implementation can be found in Section~\ref{apx:implementation} of the
appendix.\\
\textbf{Sampling:} To generate new structures, we first sample the space group
$G$ and, conditioned on $G$, the number of atoms in the asymmetric unit from the
dataset distribution. Then, starting with prior parameters $\parsnt{0}$, the
sample is generated in $n$ steps with times $t_i = i/n$ by iteratively sampling
$y$ from ${\rec(\cdot \mid \parsnt{i}; t_{i}, \alpha_{i})}$—i.e., sampling a
structure prediction $\C'$ from  ${\out(\cdot \mid \parsnt{i}, t_{i})}$ and then
$\y$ from $\sender{\cdot}{\C', \alpha_{i+1}}$—and then setting ${\parsnt{i+1} =
h(\parsnt{i}, \y)}$. The final sample $\C$ is drawn from $\out(\cdot \mid
\parsnt{n}, 1)$. $\C$ includes the vector $\mathbf{k}$ for the lattice
representation, which is multiplied with the basis matrices to obtain the
lattice $\lattice$ in its matrix form, as explained in Section
\ref{sec:symmetries}. The sample also encodes the atoms in the asymmetric unit
$\A$, their site symmetry groups $\mS$, and their positions in fractional
coordinates $\x$. The complete unit cell is reconstructed from the asymmetric
unit representation using the following procedure, based on the work of
\citet{symmcd}. First, we identify the point groups that are subgroups of $G$
and most closely match $\mS$ by minimizing the Frobenius norm of their
differences. Given these site symmetries and the predicted fractional
coordinates $\x$ for each atom in the asymmetric unit, we map $\x$ to the
closest Wyckoff positions $\x'$ using the \texttt{search\_closest\_wp} function
from the PyXtal library \cite{pyxtal}. Finally, the complete unit cell is
obtained by replicating the representatives according to their Wyckoff
positions, as implemented in the PyXtal library.
\subsection{Property-Conditioned Generation}
\label{sec:prop_gen}
The SymmBFN architecture can be adapted to generate crystal structures with
desired properties. To enable conditioning on a scalar property, we modify the
neural network to incorporate the desired value $T$ as an additional input:
$\Psi(\m_{\k}, \m_{\x}, \parsn^{\A}, \parsn^{\mS}, T, t, G)$. The target $T$ is
represented using a sinusoidal positional encoding $f_{\text{pos}}(T)$, which is
concatenated with the input features of each node. The rest of the BFN framework
operates as in the model without property conditioning. During training, the
neural network receives the target values from the dataset, while during
generation, it is provided with the desired target.

\section{Experiments}
\begin{table*}[ht]
\caption{Results on the MP-20 dataset: Proxy metrics.}
\label{tab:metrics_symm_proxy}
\vskip 0.15in
\begin{center}
\begin{small}
\begin{sc}
\centering
\begin{tabular}{ccccccccc}
\toprule
 Method & Steps & \multicolumn{2}{c}{Validity (\%) $\uparrow$} & \multicolumn{2}{c}{Coverage (\%) $\uparrow$} & \multicolumn{3}{c}{Property $\downarrow$} \\
  &   & Struct. & Comp. & Recall & Precision & wdist ($\rho$) & wdist ($N_{el}$) & jsd ($G$) \\
\midrule
DiffCSP  & 1000 & \textbf{100.0} & 83.25 & \underline{99.71} & \textbf{99.76} & 0.350 & 0.340 & 0.444 \\
\midrule
DiffCSP++ & 1000 & \underline{99.94} & 85.12 & \textbf{99.73} & 99.59 & 0.235 & 0.375 & \textbf{0.077}  \\
\midrule
SymmCD  & 1000 & 94.32 & \underline{85.85} & 99.64 & 98.87 & 0.090 & 0.399 & 0.090  \\
\midrule
FlowMM  & 500 & 96.86 & 83.24 & 99.38 &  \underline{99.63} &\textbf{ 0.075} & \textbf{0.079} & 0.545 \\
\midrule
Crystal-text-LLM & - & 99.60 & \textbf{95.40} & 85.80 & 98.90 & 0.810 & 0.440 & 0.462 \\
\midrule
\midrule
SymmBFN
& 100 & 94.27 & 83.93 & \textbf{99.73} & 99.00 & \underline{0.083} &  \underline{0.095} & \underline{0.080} \\
\bottomrule
\end{tabular}
\vskip 0.1in

\end{sc}
\end{small}
\end{center}
\vskip -0.1in
\end{table*}

\begin{table*}[ht]
\caption{Results on the MP-20 dataset: Stability and cost.}
\label{tab:metrics_symm_sun}
\vskip 0.15in
\begin{center}
\begin{small}
\begin{sc}
\centering
\begin{tabular}{ccccccc}
\toprule
 Method & Steps & Time & Stability Rate $\uparrow$ & S Cost $\downarrow$ & SUN Rate $\uparrow$ & SUN Cost $\downarrow$ \\
  & & Seconds/Sample&  (\%) & Seconds &  (\%) & Seconds \\
\midrule
DiffCSP & 1000 & 0.482 &9.9 & 4.869 & 7.5 & 6.427 \\
\midrule

DiffCSP++& 1000 & 1.573 & \textbf{13.2} & 11.917 & \textbf{9.1} & 17.286   \\
\midrule
SymmCD  & 1000 & 0.514 & 9.4 & 5.468 & 7.0 & 7.343  \\
\midrule
FlowMM  & 500 & \underline{0.275} & 9.3 & \underline{2.957} & 7.4 & \underline{3.667} \\
\midrule
Crystal-text-LLM & - & -& 6.9 & - & 5.8 & - \\
\midrule
\midrule
SymmBFN& 100 & \textbf{0.007} & \underline{11.8} &\textbf{0.059} & \underline{8.9}& \textbf{0.079} \\

\bottomrule
\end{tabular}
\end{sc}
\end{small}
\end{center}
\vskip -0.1in
\end{table*}

\textbf{Metrics:} Following prior work \cite{cdvae, diffcsp, flowmm}, we use
several metrics to benchmark our proposed model against existing approaches: a
structural and compositional validity check, the coverage recall and coverage
precision on the test set, and the Wasserstein distances between the test set
and the set of generated structures for the distributions of the density $\rho$
and the distribution of the number of unique elements in the unit cell. We
include a more detailed explanation of these metrics in
Section~\ref{apx:metrics} of the appendix. Following the work of \citet{symmcd},
we also compute the Jensen-Shannon distance between the space group distribution
of the generated structures that pass the validity checks and that of the test
set, to determine whether the models accurately capture the real-world
distribution of space groups. We classify the space groups with pymatgen's
\texttt{SpacegroupAnalyzer} \cite{pymatgen} using a tolerance of 0.1.\\
The most informative metric for de novo generation is the stability of the
generated structures. A structure is considered stable if its energy above the
convex hull — i.e., the energy difference per atom between a given structure and
the most stable structure on the convex hull for the same composition — is below
0 \cite{bartel2022review}. If the energy above hull of a structure is greater
than 0, it is likely to decompose into more energetically favourable structures.
To evaluate stability, we first relax the generated structure and determine its
energy using the CHGNet neural network potential \cite{chgnet}, and then compare
this estimate to the convex hull reported in the Materials Project \cite{ehull}.
For de novo generation, we aim to generate structures that are stable (S),
unique within the generated set of structures (U), and novel with respect to the
training dataset (N). The S.U.N. rate, a metric introduced by \citet{flowmm},
represents the proportion of generated structures that meet these criteria. To
calculate the S.U.N. rate, we first identify the stable structures as described
above and discard the unstable ones. Next, we remove duplicates within the
generated set and remove samples matching structures in the training set by
running the \texttt{StructureMatcher} from the Pymatgen library \cite{pymatgen}
with default settings. The S.U.N. rate is then defined as the ratio of the
remaining samples to the total number of generated structures. Finally, to
evaluate the model efficiency, we introduce two novel cost metrics that quantify
the average computational time required to generate a stable and S.U.N.
material, respectively. For the measurements, we generate 1,000 samples per
model using a batch size of 256 on an Nvidia RTX A5000 GPU. For all models, the
post-processing time is negligible compared to the network calls and is
therefore not included. The average time required to produce a stable structure,
i.e. the time in seconds per sample divided by the stability rate, is defined as
the S. cost. Similarly, we define the S.U.N. cost as the average time to
generate an S.U.N. structure.\\
\textbf{Dataset:} All models discussed in this work, including our proposed
method, were trained on the MP-20 dataset, a subset of the Materials Project
database \cite{jain2013commentary}. This dataset comprises 40,476 crystal
structures with up to 20 atoms per unit cell. We adopt the 60-20-20
train-validation-test split initially introduced by \citet{cdvae} and
subsequently used in all other studies we compare against. The formation energy
per atom, used for the property-conditioned generation, is computed with M3GNet
\cite{m3gnet} for each structure in the dataset.

\subsection{De Novo Generation}

\begin{figure*}[ht] 
\vskip 0.2in
\begin{center}
\centerline{\includegraphics[width=\textwidth]{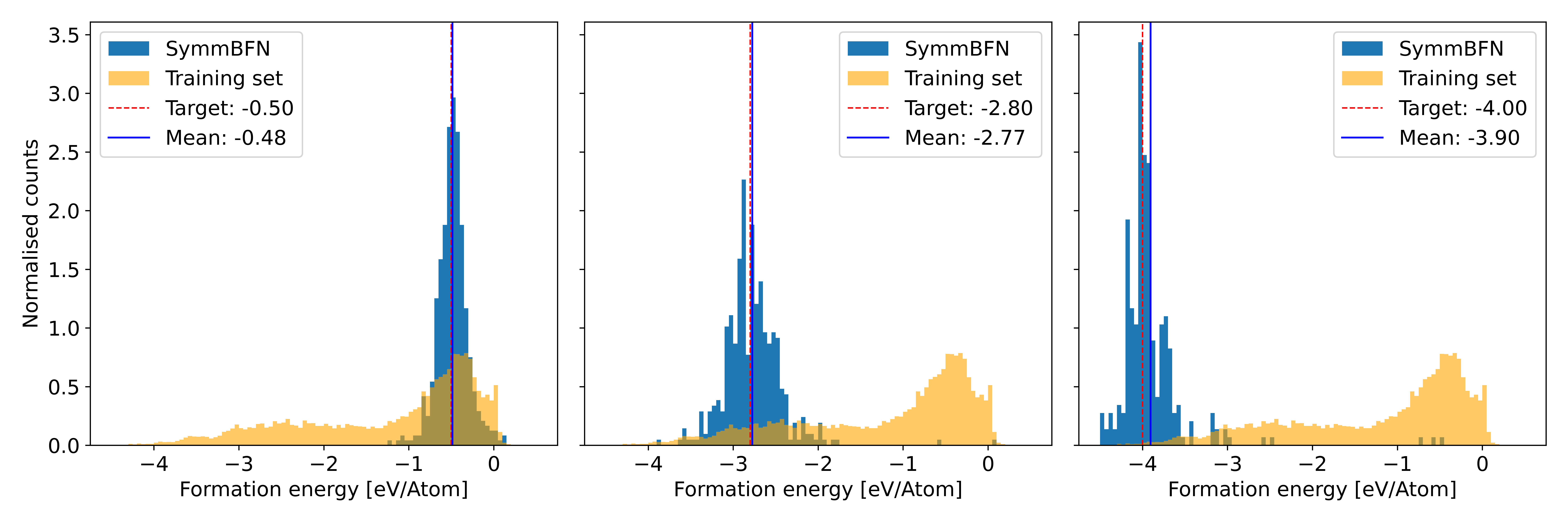}}
\caption{\textbf{Results for the property-conditioned generation for three
different target values.} The histograms show the distributions of the formation
energy per atom of the generated structures in blue and of the training set in
orange. The dashed red line represents the target of the generation while the
blue line is the mean formation energy per atom of the generated structures. }
\label{fig:eform}
\end{center}
\vskip -0.2in
\end{figure*}

In Table~\ref{tab:metrics_symm_proxy}, we benchmark SymmBFN using the metrics
described above, comparing it against several state-of-the-art models for de
novo crystal generation. These models include DiffCSP \cite{diffcsp} and FlowMM
\cite{flowmm}, both of which utilise the standard unit cell representation
(Section~\ref{sec:unitcell}), as well as DiffCSP++ \cite{diffcsp++} and SymmCD
\cite{symmcd}, which support space group-conditioned generation. Finally, we
evaluate SymmBFN in comparison to the language model-based Crystal-text-LLM
\cite{crystallm_gruver}. The Steps column reports for each method the number of
sampling steps that are used to generate a sample.  \\
SymmBFN proves to be competitive with the other generative models on all proxy
metrics. Notably, only DiffCSP++, SymmCD, and SymmBFN are capable of accurately
modelling crystal symmetries, as evidenced by the Jensen-Shannon distance
between the space group distributions of the generated structures and the test
set. This underscores the importance of incorporating crystal symmetries into
the generation process. Most importantly, SymmBFN achieves competitive results
with only 100 sampling steps, demonstrating exceptional sampling efficiency
compared to other generative models. Regarding structural validity, SymmBFN and
SymmCD perform slightly worse than the other models. We assume that this could
be because both models generate only the asymmetric unit and not the whole cell,
potentially making it more challenging to capture the arrangement of the atoms
in the entire unit cell. \\In Table~\ref{tab:metrics_symm_sun}, we present the
results of SymmBFN on the stability and cost metrics. SymmBFN ranks second in
both stability and S.U.N. rate, while its computational cost is substantially
lower than that of any other method — SymmBFN generates S.U.N. structures 50
times faster than the second-ranked model, FlowMM. The speed advantage of our
model can be attributed to the smaller computational graph, its simpler
mathematical framework, and its sampling efficiency.\\
An evaluation of the impact of varying the number of sampling steps on network
performance and additional details on the selected hyperparameter values are
provided, respectively, in Sections~\ref{apx:results} and
\ref{apx:implementation} of the appendix.

\subsection{Property-Conditioned Generation}
We evaluate our property-conditioned model using the formation energy per atom,
defined as the energy required to form a crystal structure from its constituent
elements, normalised by the number of atoms in the unit cell
\cite{jain2013commentary}. For these experiments, we use the same
hyperparameters as the model without property conditioning. During generation,
we specify three different target values for the formation energy per atom: one
at the mode, one in the tail, and one outside of the distribution of training
set values, to demonstrate the ability of the model to generate structures
across diverse target values. For each target, we generate 1,000 samples and
relax them using the CHGNet neural network. For all metastable structures
($E_{hull} \leq 0.1$), we then calculate the formation energy per atom using
M3GNet. The results, as shown in Figure~\ref{fig:eform}, demonstrate that the
model consistently generates structures with the desired formation energy per
atom. As the target value deviates further from the interval between 0 and
-1—where the training data is highly represented—the mean formation energy per
atom of the generated structures shows greater deviation from the target, along
with increased variance. Nonetheless, even for targets with sparse
representation in the training data, such as a formation energy of -4 eV/atom,
the model remains capable of proposing stable structures with the desired
property.

\section{Conclusion}
In this work, we introduced SymmBFN, a novel Bayesian flow network for the
generation of crystal structures. By explicitly incorporating the crystal
symmetries into the generation process, we were able to generate crystals more
consistent with those naturally observed. Furthermore, by allowing conditioning
on specific target properties, SymmBFN facilitates the discovery of structures
tailored to desired applications. In contrast to prior approaches based on
diffusion models, the BFN framework enables the combination of all target
variables, including the site symmetry groups and elements of the individual
atoms, into a unified framework. SymmBFN achieves competitive results for the
generation of stable and novel structures while offering a substantial speedup —
more than 50 times faster than previous generative models. This establishes BFNs
as an effective framework for crystal generation, eliminating the sampling
bottleneck of previous approaches. The demonstrated efficiency and versatility
of SymmBFN position it as a promising tool for accelerating materials design,
enabling the generation of stable, property-targeted structures with reduced
computational costs.

\section*{Software and Data}
All source code to reproduce the shown experiments will be made available on
GitHub soon. 

\newpage
\onecolumn
\bibliography{ms}
\bibliographystyle{apalike}

\newpage
\appendix
\section{Appendix}
\subsection{Metrics}
\label{apx:metrics}
For the proxy metrics, we first apply two validity metrics to the generated
structures, based on interatomic distance and charge: structural and
compositional validity. The structural validity is the ratio of generated
structures that satisfy the minimum inter-atomic distance of 0.5\AA. The
compositional validity expresses the percentage of structures with a neutral
charge, determined using SMACT \cite{smact}. To quantify the similarity of the
generated structures to the test set, we compute the coverage recall and
precision on the test set. We determine the coverage with the Euclidean distance
of the CrystalNN fingerprints \cite{crystalnnfingerprints} and normalised Magpie
fingerprints \cite{magpiefingerprints} between the generated structures and the
test set. A pair of structures is considered a match if the distance between
both fingerprints is less than the cutoffs, which are set at 10 for the Magpie
fingerprints and 0.4 for the CrystalNN fingerprints following previous work
\cite{cdvae, diffcsp, flowmm}. We further calculate the Wasserstein distances
between the test set and the set of generated structures for the distributions
of the density $\rho$ and the number of unique elements in the unit cell. To
evaluate the model, we follow the methodology of previous works to calculate the
proxy metrics: The validity and coverage metrics are calculated on 10,000
generated structures. The Wasserstein distances are evaluated on 1,000 samples
that are both structurally and compositionally valid. 

\subsection{Additional Results}
\label{apx:results}
In Table~\ref{tab:steps_proxy} and Table~\ref{tab:steps_sun}, we provide the
evaluation results of SymmBFN for different numbers of sampling steps.
Table~\ref{tab:cond_stable} shows the stability and metastability rates for the
conditioned generation.

\begin{table*}[ht]
\caption{Results on the MP-20 dataset: Proxy metrics.}
\vskip 0.15in
\begin{center}
\begin{small}
\begin{sc}
\label{tab:steps_proxy}
\centering
\begin{tabular}{ccccccccc}
\toprule
Steps & \multicolumn{2}{c}{Validity (\%) $\uparrow$} &
\multicolumn{2}{c}{Coverage (\%) $\uparrow$} & \multicolumn{3}{c}{Property
$\downarrow$} \\
   & Structural & Composition & Recall & Precision & wdist ($\rho$) & wdist ($N_{el}$) & jsd ($G$) \\
\midrule
50 & 91.74 & 83.54 & 99.76 & 98.29 & 0.163 & 0.150 & 0.092 \\
\midrule
 100 & 94.27 & 83.93 & 99.73 & 99.00 & 0.083 &  0.095 & 0.080 \\
\midrule
500 & 94.70 & 84.83 & 99.73 & 99.23 & 0.105 & 0.153 & 0.083\\
\midrule
1000 & 94.56 & 84.89 & 99.82 & 99.22 & 0.163 &0.150 & 0.089\\
\midrule
2000 & 94.90 & 84.54 & 99.76 & 99.21 & 0.179 & 0.086 & 0.081\\
\bottomrule
\end{tabular}
\end{sc}
\end{small}
\end{center}
\vskip -0.1in

\end{table*}

\begin{table*}[ht]
\caption{Results on the MP-20 dataset: Stability and cost.}
\vskip 0.15in
\begin{center}
\begin{small}
\begin{sc}
\label{tab:steps_sun}
\centering
\begin{tabular}{ccccccc}
\toprule
 Steps & Time & Stability Rate $\uparrow$ & Cost $\downarrow$ & SUN Rate $\uparrow$ & SUN Cost $\downarrow$ \\
  & Seconds/Sample&  (\%) & Seconds/Stable &  (\%) & Seconds/SUN \\

\midrule

50 & 0.004& 9.4& 0.043 & 6.0& 0.067  \\
\midrule

100 & 0.007 & 11.8 &0.059 & 8.9& 0.079 \\
\midrule

500 & 0.034 & 11.7 &  0.291 & 8.4 & 0.405  \\ 
\midrule

1000 & 0.066 & 11.8 & 0.559 & 7.3 & 0.904  \\
\midrule

2000 & 0.138 & 11.9 &  1.600 & 7.2 & 1.917  \\ 

\bottomrule
\end{tabular}
\end{sc}
\end{small}
\end{center}
\vskip -0.1in

\end{table*}

\begin{table*}[ht]
\caption{Results on the conditioned generation: Stability and metastability.}
\vskip 0.15in
\begin{center}
\begin{small}
\begin{sc}
\label{tab:cond_stable}
\centering
\begin{tabular}{ccc}
\toprule
 Target &{Stability Rate   $\uparrow$} & {Metastability Rate$\uparrow$} \\
Formation Energy& (\%)&  (\%)\\
\midrule
no target &11.8& 48.2\\
\midrule
-0.5 & 12.2 & 53.9 \\
-2.8 & 10.4 & 46.3 \\
-4.0 & 7.2 & 38.4 \\
\bottomrule
\end{tabular}
\end{sc}
\end{small}
\end{center}
\vskip -0.1in

\end{table*}

\newpage
\subsection{Implementation Details}
\label{apx:implementation}

For the neural network in the BFN, we employ the graph neural network
architecture proposed by \citet{diffcsp} based on the EGNN model
\cite{satorras2022egnn}. The input of the neural network includes the means for
the lattice representation $\m_{\k}$ and fractional coordinates $\m_{\x}$, the
parameters for the atom type distribution $\parsn^{\A}$ and symmetry
representations $\parsn^{\mS}$, the time $t$, and the spacegroup $G$. The
network ${\Psi(\m_{\k}, \m_{\x}, \parsn^{\A}, \parsn^{\mS}, t, G)}$ outputs the
scores of the output distributions after $N$ message passing layers on a
fully-connected graph. The node features are initialised with
${\bm{h}^{(i)}_{(0)} = \psi_{0}(f_{\text{atom}}(\parsn^{\A(i)}),
f_{\text{coord}}(\m_{\x}),f_{\text{sym}}(\parsn^{\mS(i)}),f_{\text{pos}}(t),
f_{\text{sg}}(G))}$, with multilayer perceptron (MLP) $\psi_{0}$, embeddings
$f(\cdot)$, and sinusoidal positional encoding $f_{\text{pos}}(t)$
\cite{Vaswani2017Attention}. The node features are then updated in each
message-passing layer as follows:
\begin{align}
 \label{eq:mpl_sym}
 &\left\{
 \begin{aligned}
     &\bm{m}^{(i,j)}_{(n)} = \psi_m(\bm{h}^{(i)}_{(n-1)}, \bm{h}^{(j)}_{(n-1)}, f_{\text{k}}(\m_{\k}), \phi_{\text{FT}}(\m_{\x}^{(j)} - \m_{\x}^{(i)}))\, , \\
     &\bm{m}^{(i)}_{(n)} = \sum_{j=1}^D \bm{m}^{(i,j)}_{(n)}\, , \\
     &\bm{h}^{(i)}_{(n)} = \bm{h}^{(i)}_{(n-1)} + \psi_h(\bm{h}^{(i)}_{(n-1)}, \bm{m}^{(i)}_{(n)})\, .
 \end{aligned}
 \right.
\end{align}

Here $\phi_{\text{FT}}(\m_{\x}^{(j)} - \m_{\x}^{(i)})$ denotes the Fourier
Transformation of the distance between the means of the fractional coordinates
for atom $j$ and $i$. $\psi_h$ and $\psi_m$ are both MLPs. We then apply four
MLPs $\psi_{k, x, a, S}$ on the node features of the last layer to obtain the
output of the network:
\begin{align}
  &\left\{
  \begin{aligned}
    &\eps_{\k} = \psi_{k}\Big(\frac{1}{D}\sum_{d=1}^D \bm{h}_{(N)}^{(d)}\Big)\, , \\
    &\eps_{\x}^{(d)} = \m_{\x} + \psi_{x}(\bm{h}_{(N)}^{(d)})\, ,  \\ 
    &\eps_{\A}^{(d)} = \psi_{a}(\bm{h}_{(N)}^{(d)})\, ,  \\
    &\eps_{\mS}^{(d)} = \psi_{S}(\bm{h}_{(N)}^{(d)})\, .
  \end{aligned}
  \right.
\end{align}

The neural network of SymmBFN has the same configuration as other generative
models \cite{diffcsp, flowmm}. The hidden dimension is 512, with an embedding
dimension of 128 and 6 message-passing layers. We train the model for 2000
epochs with a batch size of 256 until convergence., using the Adam optimiser
with default parameters and a learning rate of 0.001. We employ a
reduce-on-plateau learning rate scheduler with a factor of 0.6, patience of 100,
and a minimum learning rate of 0.0001. For the lattice and fractional
coordinates we set $\sigma=0.02$, $\beta_{\A}(1)=0.75$ for the atom types, and
$\beta_{\mS}(1)=2.0$ for the site symmetries. To balance the loss components, we
weight the individual loss functions as follows: $\lambda_{\x} = 1$,
$\lambda_{\mS} = 10$, $\lambda_{\A} = 3$, and $\lambda_{\k} = 0.1$.
\newpage
\subsection{Example Structures}
\begin{figure*}[ht] 
\vskip 0.2in
\begin{center}
\centerline{\includegraphics[width=0.6\textwidth]{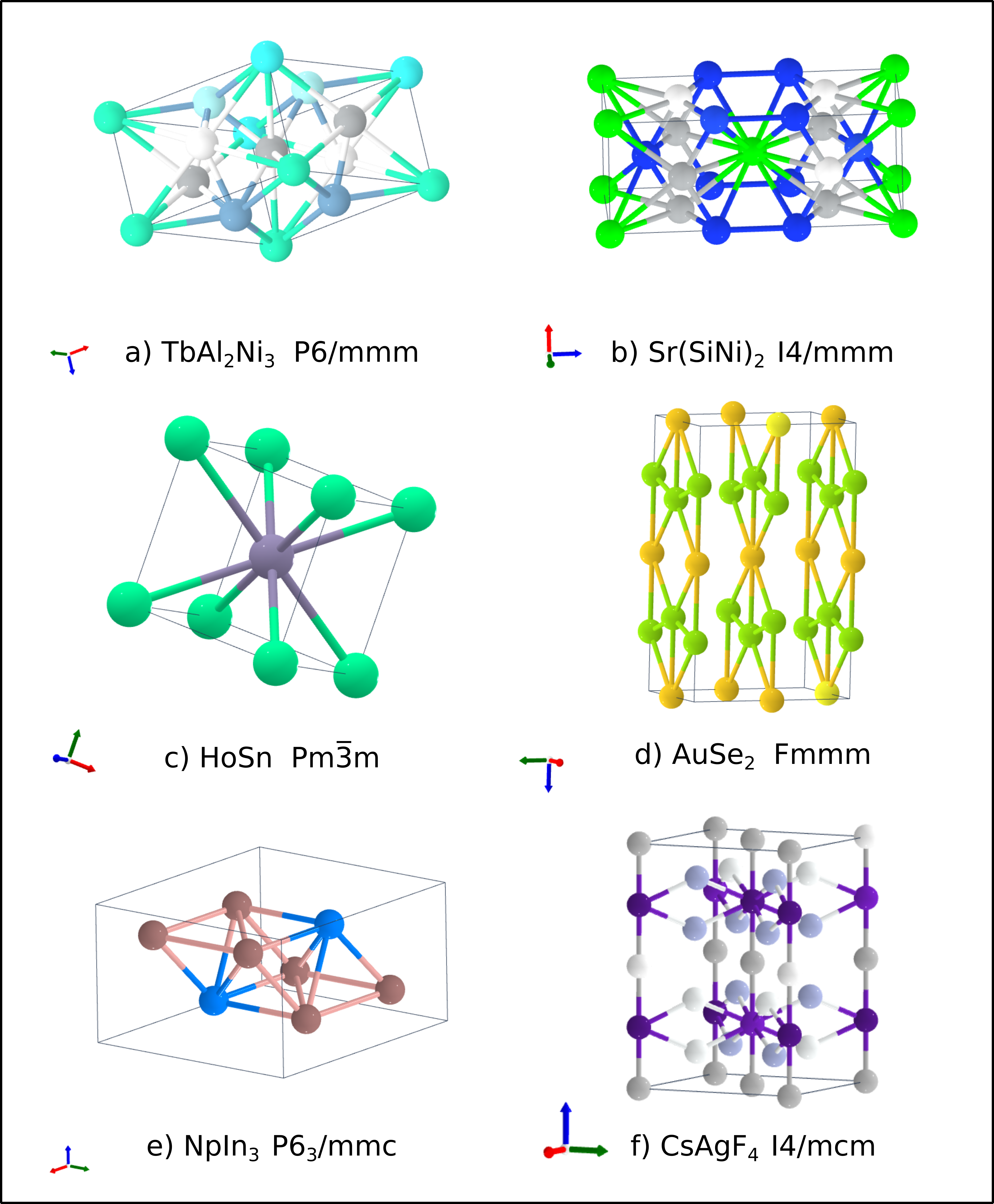}}
\caption{\textbf{Structures generated by SymmBFN with the corresponding space group.} Visualisations created with Crystal Toolkit \cite{crystaltoolkit}.}
\end{center}
\vskip -0.2in
\end{figure*}

\end{document}

%% file: math_commands.tex
\DeclareMathOperator*{\E}{\mathbb{E}}
\DeclareMathOperator{\x}{\mathbf{x}}
\DeclareMathOperator{\C}{\mathbf{c}}
\DeclareMathOperator{\X}{\mathcal{X}}
\DeclareMathOperator{\y}{\mathbf{y}}

\DeclareMathOperator{\mQ}{\mathbf{Q}}
\DeclareMathOperator{\mS}{\mathbf{S}}
\DeclareMathOperator{\mB}{\mathbf{B}}

\renewcommand{\k}{\mathbf{k}}
\DeclareMathOperator{\A}{\mathbf{a}}
\DeclareMathOperator{\lattice}{\mathbf{L}}

\DeclareMathOperator{\m}{\boldsymbol{\mu}}
\newcommand{\pt}[1]{\rho_{#1}}
\newcommand{\mt}[1]{\boldsymbol{\mu}_{#1}}
\newcommand{\kl}[2]{D_{KL}\left(#1 \parallel #2\right)}
\newcommand{\N}[2]{\mathcal{N}\left(#1 , #2\right)}

\DeclareMathOperator{\R}{\mathbb{R}}
\newcommand{\I}[1]{\boldsymbol{I}}

\newcommand{\tidx}[2]{#1_{#2}}
\newcommand{\didx}[2]{#1^{(#2)}}
\renewcommand{\vec}[1]{\boldsymbol{#1}}
\newcommand{\pars}{\theta}
\newcommand{\parsn}{\vec{\pars}}

\newcommand{\parsnt}[1]{\tidx{\parsn}{#1}}
\newcommand{\alphat}[1]{\tidx{\alpha}{#1}}
\newcommand{\yt}[1]{\tidx{\y}{#1}}
\newcommand{\constvec}[2]{\vec{#1}}
\newcommand{\0}[1]{\constvec{0}{#1}}
\newcommand{\1}[1]{\constvec{1}{#1}}

\newcommand{\Add}[1]{\didx{a}{#1}}
\newcommand{\parsdd}[1]{\didx{\pars}{#1}}
\newcommand{\oh}[2]{\mathbf{e}_{#1}}
\newcommand{\ds}[1]{\{1,#1\}}
\newcommand{\dsd}[2]{\ds{#1}^{#2}}

\newcommand{\sender}[2]{p_{_S}\left(#1 \mid #2\right)}
\newcommand{\out}{p_{_O}}

\newcommand{\rec}{p_{_R}}
\newcommand{\inp}{p_{_I}}
\newcommand{\flow}{p_{_F}}
\newcommand{\update}{p_{_U}}
\newcommand{\pred}[1]{\hat{#1}}
\newcommand{\eps}{\vec{\pred{\epsilon}}}
\newcommand{\bm}[1]{\boldsymbol{#1}}